# A new RCM mechanism for an ear and facial surgical application

Guillaume Michel[1,2], Durgesh. H. Salunkhe[1], Damien Chablat[1,3], Philippe Bordure[2]

[1]Laboratoire des Sciences du Numérique de Nantes (LS2N), UMR CNRS 6004,
1 rue de la Noë, 44321 Nantes
salunkhedurgesh@gmail.com
[2]CHU de Nantes, 5 Allée de l'Île Gloriette, 44093 Nantes
{philippe.bordure@chu-nantes.fr, guillaume.michel@chu-nantes.fr}
[3]CNRS, Laboratoire des Sciences du Numérique de Nantes,
UMR 6004 Nantes, France
damien.chablat@cnrs.fr

**Abstract.** Since the insertion area in the middle ear or in the sinus cavity is very narrow, the mobility of the endoscope is reduced to a rotation around a virtual point and a translation for the insertion of the camera. This article first presents the anatomy of these regions obtained from 3D scanning and then a mechanism based on the architecture of the agile eye coupled to a double parallelogram to create an RCM. This mechanism coupled with a positioning mechanism is used to handle an endoscope. This tool is used in parallel to the surgeon to allow him to have better rendering of the medium ear than the use of Binocular scope. The mechanism offers a wide working space without singularity whose borders are fixed by joint limits. This feature allows ergonomic positioning of the patient's head on the bed as well as for the surgeon and allows other applications such as sinus surgery.

**Keywords:** Otologic surgery, Endoscopy, RCM mechanism, spherical robot.

## 1      Introduction

In the field of ear surgery, and more broadly of microsurgery, several challenges are encountered by the surgeon. The middle ear is an anatomical entity of small volume with multiple fragile elements not to be damaged. Operations are traditionally performed under binocular loupes, which allows the surgeon to use both hands for a micro-instrument and a suction tool. More recently, the development of endoscopic otologic surgery allows better vision of hard-to-reach areas [1, 2]. However, the need of handling the endoscope limits the surgeon's capability to operate with only one instrument at a time**.** This constraint also exists in facial surgery, which has been performed with endoscopy for many years.

Currently, several robotic systems are being developed, demonstrating the interest in robotic assistance in microsurgery. Prof. Sterkers' team [3] has developed a six-degree-of-freedom tele-operated robot that can be used with the operating microscope. The design was not originally intended as an endoscopic surgical aid, and its manipulation requires the use of one hand. Other systems allow a precise part of the surgery, programmed by the surgeon and performed by the robot, to be performed: insertion of an electrode into the cochlea [4], milling of a mastoidectomy [5], etc ... At the level of the facial mass, there is currently no robotic system to assist the surgeon in the procedure.

The problem is therefore to improve the safety of the gesture in this high-risk environment, by assisting the surgeon mainly in the use of endoscopy. The objective of this work is to design a robotic system to assist the surgeon as a third hand, holding the endoscope and following the surgeon's gestures. The objective of the work presented in this paper is to introduce the workspace, the variations in operating positions, and the advantages of an "agile eye" type robotic system coupled with a Remote Center Motion (RCM) mechanism in this context.

## 2  Workspace characterization

During ear or facial surgery, the patient is positioned supine. However, the precise position of the head varies: depending on the type of table and headrest, on the patient's morphology, on the type of surgery, and according to the surgeon's practice. Indeed, during sinus surgery, the head is most often oriented in anterior flexion, in order to have easier access to the antero-superior spaces of the paranasal sinuses, such as the anterior ethmoid and the naso-frontal canal. Conversely, during stapes surgery for otosclerosis, the head is positioned in hyperextension in order to facilitate access to this anatomical region.

The choice of the robot architecture is important to allow the robot to adapt to these different situations. Larger the workspace, thanks to high amplitudes of movements, less will the surgeon be constrained by the robot. Indeed, the ear must not be positioned according to the robot, but the robot must adapt to the different positions and morphologies. A choice of architecture based on the middle ear could strongly constrain its field of action to extend the use of the robot to other applications such as sinus surgery or neurosurgery. In other words, a robot that is too optimized for the middle ear, for example, could constrain the robot in other operations. The robot must allow a better vision of the area operated on by the surgeon than with a microscope without reducing the number of tools used and guaranteeing its freedom of movement (Figure 1).

For this reason, a post-scan study was performed in patients of different ages and sexes, in the outer and middle ear and in the facial region. However, the architecture that would be chosen must not only allow an adaptation to this anatomical environment but also to peri-anatomical position variations. The ear workspace consists of the middle ear box and the external ear canal. Endoscopy is most often used for the treatment of pathologies affecting these two anatomical zones. It is possible to use the endoscope for mastoid surgeries, as far as for internal auditory canal; but the space is larger and variable according to the drilling performed by the surgeon, therefore less constraining. This working space has several particularities: first, its size is physiologically variable, depending on the subjects. However, for non-pathological cases, there are no significant differences according to sex, age or side of the ear [6]. Secondarily, in pathology, it can vary from a complete absence of these zones (aplasia) to a volume extended at will by the surgeon (in carcinology for example).

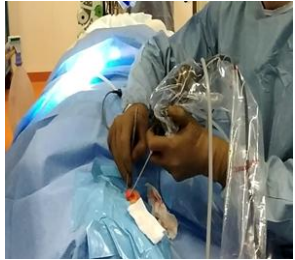
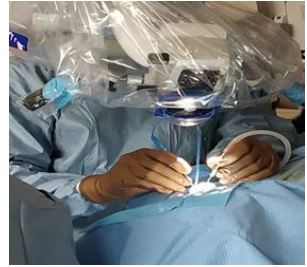

(a) The use of 1 hand to hold endoscope limits the number of instruments

(b) The surgeon can use 2 instruments while using a microscope

Figure 1 : The comparison of the number of instruments possible to use simultaneously while using an endoscope and a microscope.

In the literature, the analyses are most often radiological and concern the external auditory canal, the ossicles or the mastoid. Thus, the parisian team developing Robotol [3] used 12 scanners to measure a middle cylinder corresponding to the external auditory canal and the visible part of the middle ear. Dillon [5], on cadaveric models, and Cros [7], from 10 or more scans, were interested in the mastoid only; but the mastoid is not the preferred working area for endoscopic surgery, and can be enlarged on request by drilling. Pacholke [8] found an average middle ear volume of 0.58 cm$^3$ from 15 scans, with a maximum axial dimension of 1.57 cm, while Mas [9] evaluated it at between 5.25 and 6.22 cm$^3$ from 18 scans.

The largest study found from 100 scans [6] evaluated the volume of the external auditory canal at 1.4 mL and that of the middle ear at 1.1 mL. This volume decreases significantly in the presence of chronic otitis media. In total, the data are highly variable across studies, and there is no geometric measurement of the middle ear. It is thus of paramount importance to make a geometric atlas in order to better define our workspace. This study is based on scans of petrous bones from a population of variable age and sex (n=16, patients from 2 to 79 years old). Measurements were taken on the three axes of the tympanic body, from the hypotympanum to the attic, but also from the external auditory canal, to the bony canal-fibro-cartilaginous junction, and at the sulcus level. The mean measurements as well as the extreme values are shown in Figure 2. These values assist in evaluating the maximum span of endoscope in ontological surgery as shown in the same figure. The totality of the data is presented in Table 1.

In the facial region, the workspace covers the nasal fossae from the floor to the roof of the ethmoid, the maxillary sinuses to their lateral edge, and the posterior pharyngeal wall at the posterior border. The volumes of the different sinuses have been extensively studied in the past, as well as the influence of different pathologies, infec-tious or malformative on their size or growth. The maxillary sinus has been the most studied [10, 11], and shows, for example, a decrease in volume with age and loss of maxillary teeth [12]. The other sinuses are also studied in terms of size and anatomical ratios, such as the sphenoid sinus [13]. However, we have not found any study that looked at the dimensions of the paranasal sinuses as a whole, setting upper and lower limits to define a robotic workspace.

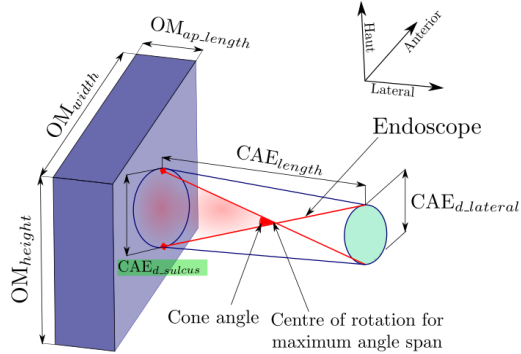

Figure 2 : schematic workspace of the external (cylinder) and middle ear

**Table 1**: Data from petrous bone scan analysis on 16 patients
(CAE: External Auditory Canal, OM: Middle Ear).

| Age | CAE diameter lateral extremity | CAE diameter at sulcus | CAE length | OM Height | OM Width | OM Anteroposterior-posterior length |
|---|---|---|---|---|---|---|
| 50 | 7,2 | 9,1 | 26,9 | 19,4 | 11,7 | 5,1 |
| 79 | 5,3 | 11,1 | 34,4 | 15,3 | 12,4 | 6,5 |
| 2 | 4,0 | 9,2 | 22,5 | 14,8 | 11,4 | 6,2 |
| 58 | 3,9 | 6,3 | 28,5 | 15,3 | 8,8 | 6,8 |
| 59 | 7,1 | 9,0 | 35,3 | 14,5 | 12,1 | 4,4 |
| 37 | 6,8 | 8,4 | 25,3 | 18,5 | 11,2 | 5,5 |
| 71 | 6,5 | 10,2 | 27,4 | 15,5 | 11,2 | 5,2 |
| 4 | 4,8 | 8,8 | 23,2 | 15,1 | 12,3 | 6,8 |
| 51 | 4,3 | 6,5 | 26,9 | 14,1 | 9,1 | 7,2 |
| 44 | 7,3 | 9,3 | 31,3 | 19,1 | 10,6 | 4,9 |
| 29 | 6,6 | 6,3 | 28,1 | 15,3 | 9,2 | 4,7 |
| 32 | 6,8 | 6,2 | 23,1 | 14,5 | 7,6 | 4,9 |
| 18 | 5,6 | 6,6 | 25,2 | 14,5 | 10,8 | 6,8 |
| 5 | 5,5 | 7,5 | 27,2 | 16,2 | 11,6 | 6,2 |
| 8 | 5,1 | 6,9 | 22,2 | 16,6 | 8,4 | 4,1 |
| 78 | 5,7 | 6,6 | 26,6 | 15,3 | 8,5 | 3,1 |
| **Average** | **5,79** | **7,85** | **26,75** | **17,35** | **10,10** | **4,10** |
| **σ** | **1,15** | **1,58** | **3,87** | **1,68** | **1,57** | **1,17** |

We have therefore carried out a study on scanners of the paranasal sinuses, in a population of variable age and sex (n=23, patients from 11 to 95 years old). Measurements include (i) the distance between the piriform orifice and the posterior pharyngeal wall, (ii) the lateral wall of the maxillary sinus and nasal septum on each side, (iii) the distance between the floor of the nasal fossae and the roof of the ethmoid at the level of the naso-frontal canal, (iv) the distance between the nasal septum and the middle meatus on each side, and (v) the height of the piriform orifice, which would correspond to the endoscope entrance orifice, the probable site of the MCR. The mean and extreme values are shown in Figure 3. The data related to nasal measurements are presented in Table 2.

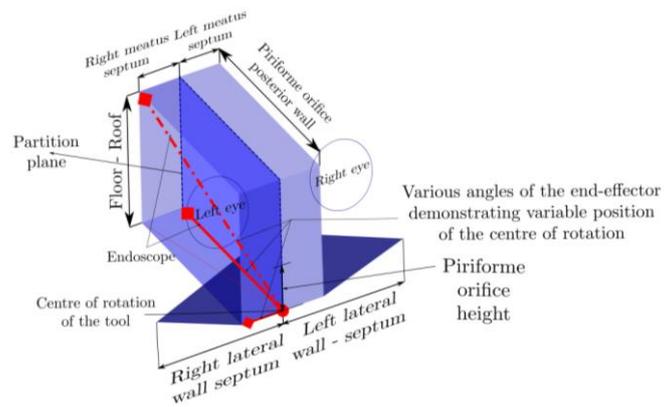

Figure 3: schematic workspace of paranasal sinuses, coronal view: maxillary sinuses (triangles) and vertical workspace from the floor of the nasal cavities to the roof of the ethmoid

**Table 2**: Data from paranasal sinuses scan analysis on 23 patients

|  | Age | Piriforme orifice – Posterior wall | Right lateral wall - septum | Left lateral wall - septum | Floor - Roof | Right meatus - septum | Left meatus - septum | Piriforme orifice height |
|---|---|---|---|---|---|---|---|---|
|  | 26 | 73 | 32 | 43 | 54 | 13 | 14 | 26 |
|  | 40 | 79 | 37 | 40 | 57 | 14 | 15 | 28 |
|  | 57 | 59 | 36 | 35 | 47 | 12 | 17 | 29 |
|  | 11 | 77 | 39 | 34 | 42 | 10 | 11 | 21 |
|  | 31 | 85 | 42 | 43 | 66 | 16 | 12 | 31 |
|  | 73 | 78 | 44 | 45 | 61 | 14 | 14 | 35 |
|  | 54 | 94 | 44 | 36 | 52 | 17 | 12 | 30 |
|  | 95 | 87 | 44 | 37 | 51 | 13 | 17 | 29 |
|  | 32 | 84 | 41 | 39 | 52 | 12 | 16 | 31 |
|  | 62 | 74 | 41 | 30 | 48 | 11 | 14 | 29 |
|  | 51 | 67 | 41 | 37 | 55 | 11 | 17 | 31 |
|  | 95 | 83 | 38 | 40 | 52 | 12 | 18 | 34 |
|  | 50 | 83 | 45 | 41 | 56 | 13 | 13 | 27 |
|  | 55 | 63 | 43 | 47 | 53 | 17 | 13 | 32 |
|  | 52 | 82 | 44 | 43 | 67 | 15 | 13 | 36 |
|  | 81 | 79 | 27 | 27 | 49 | 18 | 18 | 29 |
|  | 80 | 76 | 36 | 44 | 61 | 15 | 16 | 34 |

|         | 83    | 67    | 38    | 40    | 60    | 10    | 9     | 34    |
|---------|-------|-------|-------|-------|-------|-------|-------|-------|
|         | 67    | 72    | 35    | 37    | 50    | 13    | 16    | 26    |
|         | 29    | 81    | 42    | 44    | 67    | 11    | 15    | 24    |
|         | 76    | 67    | 42    | 41    | 53    | 11    | 13    | 29    |
|         | 47    | 83    | 35    | 39    | 59    | 14    | 10    | 26    |
|         | 48    | 79    | 36    | 42    | 62    | 13    | 14    | 29    |
| Average | 56,3  | 77,04 | 39,22 | 39,30 | 55,39 | 13,26 | 14,22 | 29,57 |
| σ       | 22,52 | 8,35  | 4,51  | 4,78  | 6,63  | 2,26  | 2,47  | 3,67  |

It can be noted that the maximum travel of the endoscope can reach up to 90 degrees thus emphasizing the importance of having feasible workspace with at least ±45 degrees. In some cases, the partition separating the two nasal cavities is removed resulting in an enlarged workspace.

## 3 Remote Center Motion and Spherical Wrist

### 3.1 Mechanisms with RCM

A remote fixed point, with no physical revolute joint over there, around which a mechanism or part of it can rotate is called remote center of motion [14]. Most RCM mechanisms used for medical applications have parallelograms and two rotary actuators in series for rotational movements as in Figure 4 [16, 17]. Two bevel gear are used to transmit the motion from the base to the second joint. However, either this transmission has operating clearances or a preload that produces incompatible friction that makes the transmission non-reversible. A similar system can also be completely passive to make motion acquisition of the surgeons as in Figure 5 [18]. The objective of this article is to associate the following parallel mechanism to the designs presented.

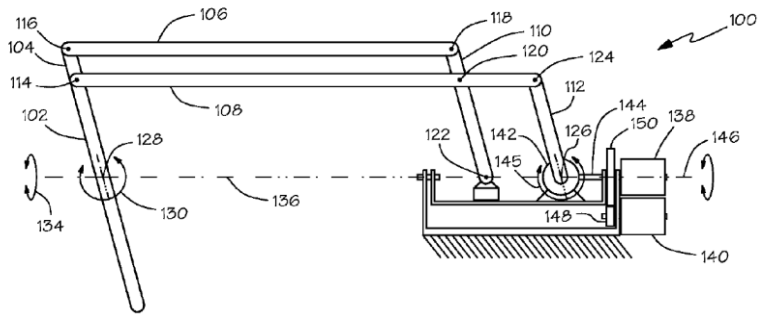

Figure 4 : Robotic manipulator with remote center of motion and compact drive [16]

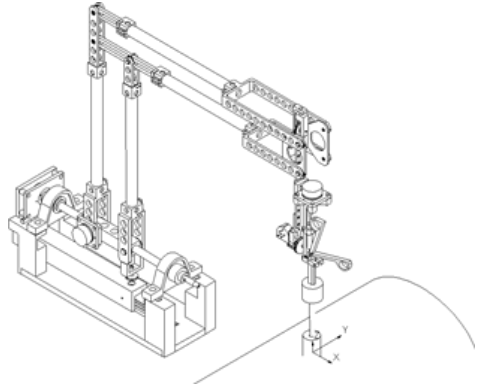

Figure 5 : BlueDRAGON mechanism and its coordinate system [18]

### 3.2 Properties of the Agile Eye

The agile eye was designed to move a camera very fast with three rotational motions [19]. A second, less-known version has only two degrees of rotational freedom [20]. Two serial chains consisting of revolute joints whose axes intersect at a single point are connected to constrain the orientation of the V vector as shown in Figure 6. A modification of this mechanism has been presented in [21, 22] to be an effector of a 5-axis machine tool.

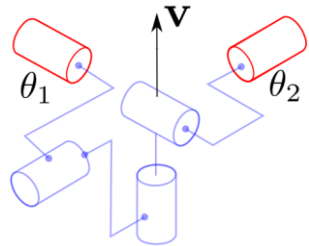

Figure 6 : Two DOF spherical mechanism, the Agile Eye [19]

The direction of end-effector, called **V**, defined by two rotations around axis **x** and **y** are

$$V = \begin{bmatrix} \sin(\beta) & -\sin(\alpha)\cos(\beta) & \cos(\alpha)\cos(\beta) \end{bmatrix}^T$$

The inverse kinematic model can be solve

$$\theta_1 = -\arctan\left(-\sin(\alpha)/\cos(\alpha)\right), \quad \theta_2 = -\arctan\left(-\sin(\beta)/\cos(\alpha)\cos(\beta)\right)$$

The main advantage of this mechanism is that singular configurations are when $\theta_1 = \pm\pi$ or $\theta_2 = \pm\pi$. However, during a practical implementation of this robot, the internal collisions limit the movements. By controlling the inclination of the parallel-ogram with $\theta_2$ and its orientation with $\theta_1$, we obtain the mechanism of Fig. 7.

The orientation of the endoscope carried by the mechanism, defined by V, is identical to that defined by the agile eye thanks to the double parallelogram. Figures 6 and 7 show the agile eye in its isotropic configuration, which is also the optimal configuration for parallelograms (the posture furthest from the constraint singularities). This mechanism allows a large deflection without internal collision and without singularities as shown in Figure 8. A patented mechanism [23], not shown in the figure, ensures patient safety during operations, quick cleaning of the endoscope optics and rotation if the optics are inclined.

To position the robot in relation to the patient's ear or sinuses, a Cartesian displacement mechanism can add the necessary translational mobility. It can be on a mobile base [3] or fixed on the bed using the fixation rails as shown in Figure 9.

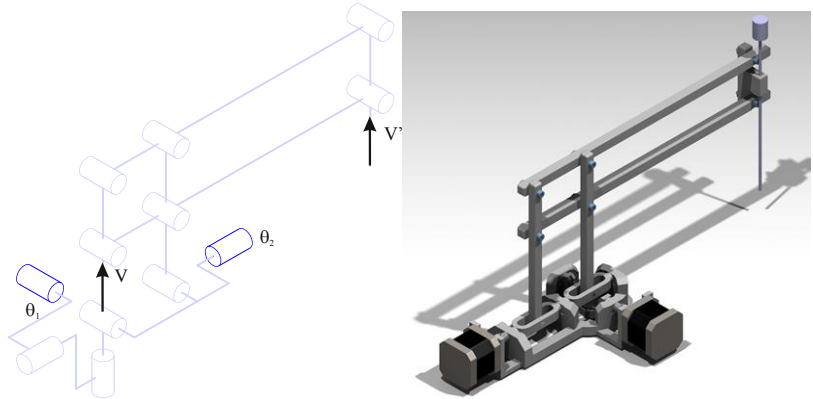

Figure 7 : Two DOF spherical mechanism coupled with a double parallelogram

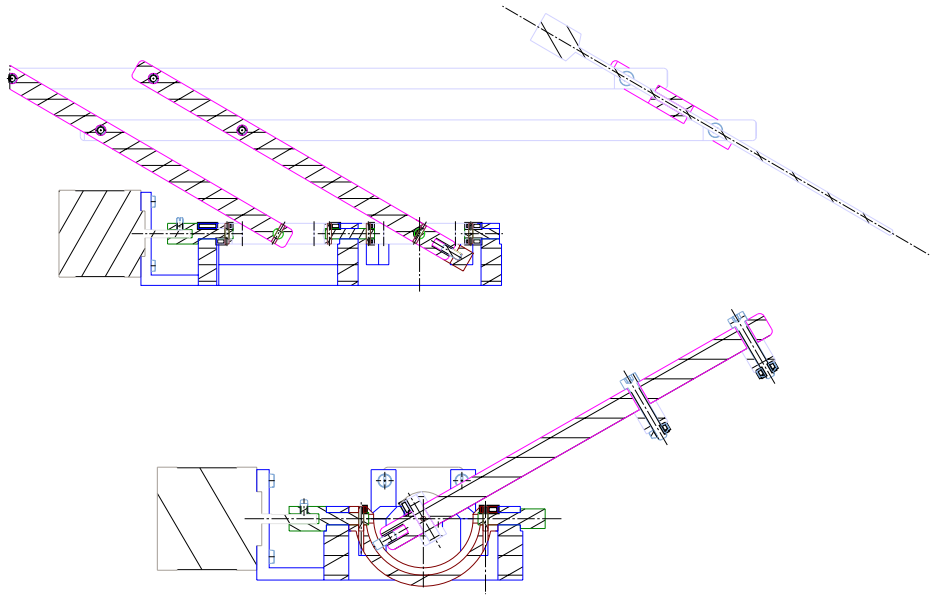

Figure 8 : Maximum angles of rotation of the parallelograms and the agile eye

## 4　　Conclusions

In this article, an analysis of surgeons' needs for ear and sinus operations is presented. A new mechanism with two degrees of rotational mobility is presented. This mechanism combines a spherical mechanism and a double parallelogram to perform an RCM to carry an endoscope. The result is a large working area with no singularity for a very compact mechanical design. The height of the parallelogram as well as the translational displacements of the Cartesian robot will be optimized to respond to variations in patient anatomy. A prototype is under construction to validate its mobility and its use for automatic tool tracking with the endoscope. A safety device must also allow the endoscope to be ejected if the patient wakes up, rapid cleaning of the optic and rotation along the insertion axis if the optic is tilted.

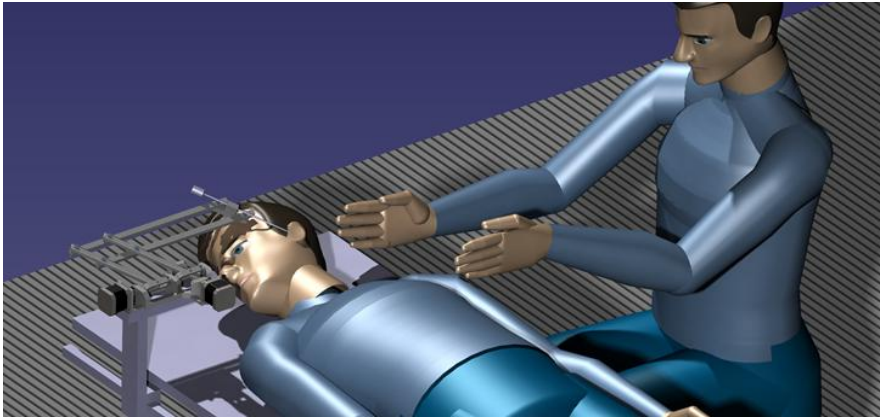

Figure 9 : Positioning of the robot with a Cartesian mechanism

## Acknowledgment

The project is receiving financial support from the NExT (Nantes Excellence Trajectory for Health and Engineering) Initiative and the Human Factors for Medical Technologies (FAME) research cluster.